\PassOptionsToPackage{table}{xcolor}

\documentclass{article} 
\usepackage{iclr2027_conference,times}

\usepackage{booktabs}
\usepackage{multicol}
\usepackage{multirow}
\usepackage{amsmath,amsfonts,bm}
\usepackage{xspace}

\usepackage[table,dvipsnames]{xcolor}

\usepackage{microtype}
\usepackage{cuted}
\usepackage{enumitem}
\usepackage{pifont}

\usepackage[most]{tcolorbox}
\usepackage{titletoc}
\usepackage{url}
\usepackage{subcaption}
\usepackage{tcolorbox}
\renewcommand{\arraystretch}{1.2}

\usepackage[
    colorlinks=true,
    citecolor=BlueViolet,
    linkcolor=purple,
    urlcolor=BlueViolet
]{hyperref}

\newcommand\DoToC{%
  \startcontents
  \printcontents{}{1}{\noindent \textbf{\Large{Table of Contents in Appendix}}\vskip3pt\vskip5pt}
  \vskip3pt\vskip5pt
}

\usepackage{amsmath,amsfonts,bm}

\def\eqref#1{equation~\ref{#1}}

\def\1{\bm{1}}

\DeclareMathAlphabet{\mathsfit}{\encodingdefault}{\sfdefault}{m}{sl}
\SetMathAlphabet{\mathsfit}{bold}{\encodingdefault}{\sfdefault}{bx}{n}

\usepackage{url}

\usepackage[ruled]{algorithm2e} 
\usepackage{booktabs}       
\usepackage{graphicx}       
\usepackage{amsfonts}       
\usepackage{nicefrac}       
\usepackage{microtype}      
\usepackage{xcolor}         
\definecolor{best}{RGB}{255,218,218}
\definecolor{second}{RGB}{255,232,190}
\definecolor{third}{RGB}{255,249,190}
\usepackage{xspace}
\usepackage{amsmath}
\usepackage{amsthm}
\usepackage{multicol}
\usepackage{multirow}
\usepackage{enumitem}
\usepackage{color}
\usepackage{subcaption}
\usepackage[most]{tcolorbox}
\usepackage{wrapfig}

\SetAlFnt{\small}
\SetAlCapFnt{\small}
\SetAlCapNameFnt{\small}
\SetAlCapHSkip{0pt}

\newcommand{\ourmethod}{{{4D-HOF}}\xspace}
\newcommand\mypara[1]{\vspace{1mm}\noindent\textbf{#1}}

\title{\ourmethod: {H}and-{O}bject {F}low Matching for \\ Feed-Forward 4D Interaction Reconstruction}

\author{%
  Shiqi Li$^{1}$\
  \quad
  Sean Cho$^{1}$\
  \quad
  Yijie Li$^{1}$\
  \quad
  Fengzhi Guo$^{1}$\
  \quad 
  Bowen Wen$^{2}$
  \quad 
  Cheng Zhang$^{1}$ \\ [3pt]
$^1$Texas A\&M University \quad 
$^2$NVIDIA \\ [3pt]
\href{https://tamu-visual-ai.github.io/4D-HOF}{https://tamu-visual-ai.github.io/4D-HOF}
}

\iclrfinalcopy 
\begin{document}

\maketitle
\lhead{}

\begin{figure}[h]
\centering
\includegraphics[width=1\textwidth]{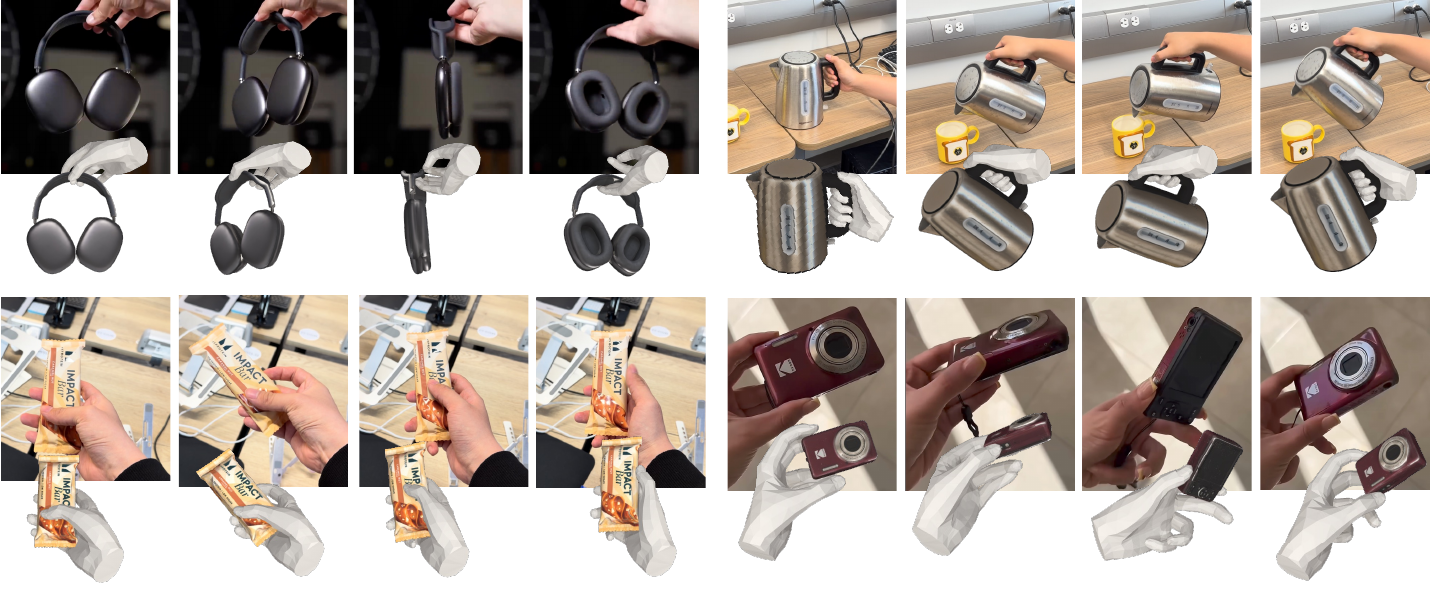}
\caption{\small \textbf{4D Hand-Object Flow Matching (\ourmethod)}. Given a monocular video of a hand interacting with an unseen object, we learn a flow matching model for accurate and efficient 4D hand-object interaction reconstruction. Our method generalizes across diverse and challenging scenarios, including ambiguous depth (headset), occlusion (kettle), complex articulated hand poses (bar), and large object rotation (camera). Moreover, \ourmethod is substantially faster than prior optimization-based methods due to our feed-forward design.}
\label{fig:teaser}
\vspace{3mm}
\end{figure}

\begin{abstract}
Existing methods for 4D hand-object reconstruction often rely on costly per-sequence optimization, while generative approaches typically synthesize interactions from random noise, which can lead to unstable interaction prediction. 
We introduce \ourmethod, a feed-forward framework that reconstructs 4D hand-object interactions from coarse but informative estimates produced by vision foundation models. 
Concretely, we learn a conditional flow matching model that transports foundation-model-derived hand-object states toward an interaction manifold, allowing the model to correct errors in translation, rotation, and alignment in a feed-forward manner.
A key advantage of our generative formulation is that it naturally enables test-time guidance within the transport process. Rather than applying a separate post-hoc optimization after reconstruction, we directly steer the evolving generative states using physical interaction constraints and observed 2D evidence, allowing the reconstruction to be refined as part of the generative process itself.
By training the generative model on diverse datasets, \ourmethod generalizes robustly to challenging in-the-wild scenarios. Experiments on out-of-domain benchmarks show that \ourmethod achieves state-of-the-art performance, producing more stable and accurate 4D hand-object reconstructions.
\end{abstract}

\section{Introduction}

Reconstructing 4D hand-object interaction (HOI) from monocular video is a fundamental problem for understanding human behavior and enabling a variety of applications such as robot learning~\citep{he2025dexvlg, zhong2025dexgrasp, paliwal2026dexterous}, extended reality~\citep{marchand2015pose}, and human-computer interaction~\citep{pavlovic2002visual}. Despite recent progress~\citep{fan2024hold, wang2025magichoi, shi2026agile, xu2026choir}, recovering accurate hand-object geometry and interaction remains challenging due to diverse object geometry and pose, severe occlusion, and the inherent ambiguity of monocular observations.

To address these challenges, existing approaches mainly fall into two lines. Optimization-based methods~\citep{fan2024hold, wang2025magichoi, shi2026agile, aytekin2025follow, zhu2025reconstructing} formulate HOI reconstruction as a per-sequence fitting problem, enforcing interaction constraints through manually designed objectives such as contact, rendering, and interaction stability. While effective, these methods are computationally expensive and often brittle under imperfect initialization or occlusion. 
Generative methods~\citep{ye2023diffusion, ye2024g, christen2024diffh2o, li2025latenthoi, ye2026whole} learn priors over HOI distributions from large-scale data, yet often start from random noise or rely on category-level priors with closed-set vocabularies, making it difficult to preserve instance-specific geometry and pose cues. 
This raises a key question: can we retain and learned priors of generative modeling while starting from informative observations rather than random noise?

Vision foundation models provide a natural starting point. Recent large-scale pretrained models can estimate hand pose, object geometry, depth, and scene structure from monocular inputs~\citep{carion2026sam, oquab2023dinov2, piccinelli2025unidepthv2, lin2025depth, wen2024foundationpose, pavlakos2024reconstructing, potamias2025wilor}. These models produce coarse yet informative estimates that capture rich geometric cues. Indeed, recent works have begun to incorporate foundation models into HOI reconstruction ~\citep{liu2025easyhoi, wang2026arthoi, lin2025zero, xu2026choir}. However, they typically treat these estimates as initialization and rely on downstream heuristic optimization to enforce consistency, which reintroduces inefficiency and instability.

In this work, we instead couple foundation-model perception directly with generative modeling. We use them to define the source distribution of a conditional generative transport process. Starting from coarse but informative hand-object states, we learn how these states should evolve toward physically plausible interactions.
Specifically, we adopt flow matching and stochastic interpolates~\citep{liu2022let, shi2023diffusion, albergo2025stochastic} to construct a generative bridge that transports the initial estimates toward an HOI manifold. To ground refinement in observed evidence and exploit interaction dynamics, the bridge is conditioned on both the input video frames and the temporal context of the initial 4D hand-object sequence. This formulation enables unified correction of geometry, pose, and alignment in a feed-forward manner, while preserving instance-specific cues from the input. 
Importantly, the generative formulation also gives us a natural mechanism for test-time control to further enhance fine-grained physical plausibility and consistency with the 2D observation.
Our method, termed \textbf{\ourmethod}, is a feed-forward framework that takes foundation-model estimates as input and jointly refines hand-object interactions. We summarize our contributions as follows:
\begin{itemize}[leftmargin=1.3em,topsep=2pt,itemsep=1pt,parsep=0pt,before=\vspace{-6pt}]
    \item We formulate 4D HOI reconstruction as conditional generative transport from foundation-model-derived hand-object state initialization to physically plausible interactions, avoiding random noise as initialization to flow matching and costly per-sequence optimization.
    \item We design a conditional flow matching model that jointly refines hand-object geometry, pose, and alignment using visual, geometric, and temporal cues.
    \item We introduce test-time guidance enabled by the generative transport process, which steers evolving states with image, interaction, and temporal constraints without separate post-hoc optimization.
    \item Experiments on 4D HOI benchmarks and challenging in-the-wild scenarios demonstrate state-of-the-art performance in hand estimation, object reconstruction, interaction consistency, and temporal stability, while running over 5 times faster than prior methods.
\end{itemize}
\section{Related Work}

\textbf{Hand-Object Reconstruction.}
Early image-based methods jointly regress hand and object meshes using parametric hand priors and contact constraints~\citep{hasson2019learning, ye2022s}. Video-based methods typically optimize each sequence with hand-crafted geometric and interaction objectives~\citep{hasson2021towards, wen2023bundlesdf, fan2024hold, wang2025magichoi, lin2025zero, shi2026agile, aytekin2026grasp, on2025bigs, liu2025generalizable, aboukhadra2026ghost}, incurring high computational costs and sensitivity to initialization. Feed-forward approaches accelerate inference~\citep{chen2026forehoi} but emphasize geometry without explicitly modeling interaction dynamics. Our method instead learns an interaction prior that maps coarse estimates to physically plausible interactions for feed-forward reconstruction.

\textbf{Foundation Models for 3D Understanding.}
Foundation models provide strong priors for segmentation~\citep{carion2026sam}, visual representations~\citep{oquab2023dinov2}, monocular and multi-view geometry~\citep{kong2026moge, piccinelli2025unidepthv2, lin2025depth, wen2025foundationstereo, wen2026fast, wang2025vggt}, object pose~\citep{wen2024foundationpose}, hand reconstruction~\citep{pavlakos2024reconstructing, potamias2025wilor}, and 3D generation~\citep{chen2026sam, xiang2026native}. Recent HOI reconstruction pipelines use these models for initialization~\citep{wang2026arthoi, lin2025zero, xie2026cari4d}, but independently estimated components often have inconsistent scale, depth, and orientation. Existing methods resolve these discrepancies through additional optimization~\citep{shi2026agile, aytekin2026grasp}; we learn to jointly refine the estimates without per-sequence optimization.

\textbf{Generative Priors for Hand-Object Interaction.}
Generative models capture HOI distributions through diffusion~\citep{ho2020denoising, christen2024diffh2o, li2025latenthoi} or language modeling~\citep{huang2025hoigpt}, primarily for interaction synthesis. Complementary denoising methods learn spatio-temporal priors to refine noisy HOI trajectories~\citep{zhou2022toch, liu2024geneoh, luo2024physics}, demonstrating the potential of learned priors for HOI correction. For reconstruction, Diff-HOI~\citep{ye2023diffusion} and G-HOP~\citep{ye2024g} incorporate generative priors, but they rely on noise-driven generation or category-level priors with closed-set vocabulary. Most closely related, CHOIR~\citep{xu2026choir} learns a flow-matching prior to correct relative hand-object 1D depth displacement along the viewing ray from initializations, while contact-aware optimization refines the full interaction. Building on flow matching and stochastic interpolants~\citep{liu2022let, shi2023diffusion, albergo2025stochastic}, we instead learn a conditional bridge that jointly refines entire HOI states, making the learned prior the primary reconstruction mechanism, complemented by test-time guidance.

\textbf{Agentic Real-to-Sim.} Recent advances in agentic AI motivate its integration into 3D reconstruction pipelines. AgentSTAR~\citep{mazur2026agenticstar} couples visual reasoning with numerical optimization in a render-and-compare loop to recover object geometry, articulation, and motion, but does not explicitly model hand–object interactions. Inspired by GPT-6 Astra’s capabilities, we conduct pilot experiments on agentic HOI reconstruction, detailed in the Appendix~\ref{ss:gpt6}. Our preliminary findings suggest that this approach is promising but incurs substantial runtime and costs. Our proposed feed-forward method offers an efficient reconstruction module that agents could use to accelerate this process.

\section{Approach}

\subsection{Problem Formulation} \label{ss:problem}
\textbf{4D HOI reconstruction.}
Given an $N$-frame monocular RGB video $\mathcal{I} = \{\mathbf{I}^i\}_{i=1}^N$, where a hand interacts with an object of arbitrary shape, our goal is to reconstruct the underlying 4D hand-object interaction. We aim to reconstruct the object mesh $\mathbf{m}_o$ with corresponding per-frame 6-DoF poses $\mathbf{o}^i = (\mathbf{R}^{o}, \mathbf{t}^{o})$, where $\mathbf{R}^{o} \in SO (3)$ and translation $\mathbf{t}^{o} \in \mathbb{R}^3$. For the hand, we estimate the per-frame MANO~\citep{romero2022embodied} parameters, consisting of shape $\bm{\beta}$ and pose $\mathbf{h}^i = (\mathbf{R}^{h}, \mathbf{t}^{h})$. We denote the full interaction trajectory as $\mathcal{S} = \{\mathbf{s}^i\}_{i=1}^N$, where $\mathbf{s}^i = [\mathbf{h}^i, \mathbf{o}^i]$.

\begin{figure}[t]
    \centering
    \small
    \includegraphics[width=1\textwidth]{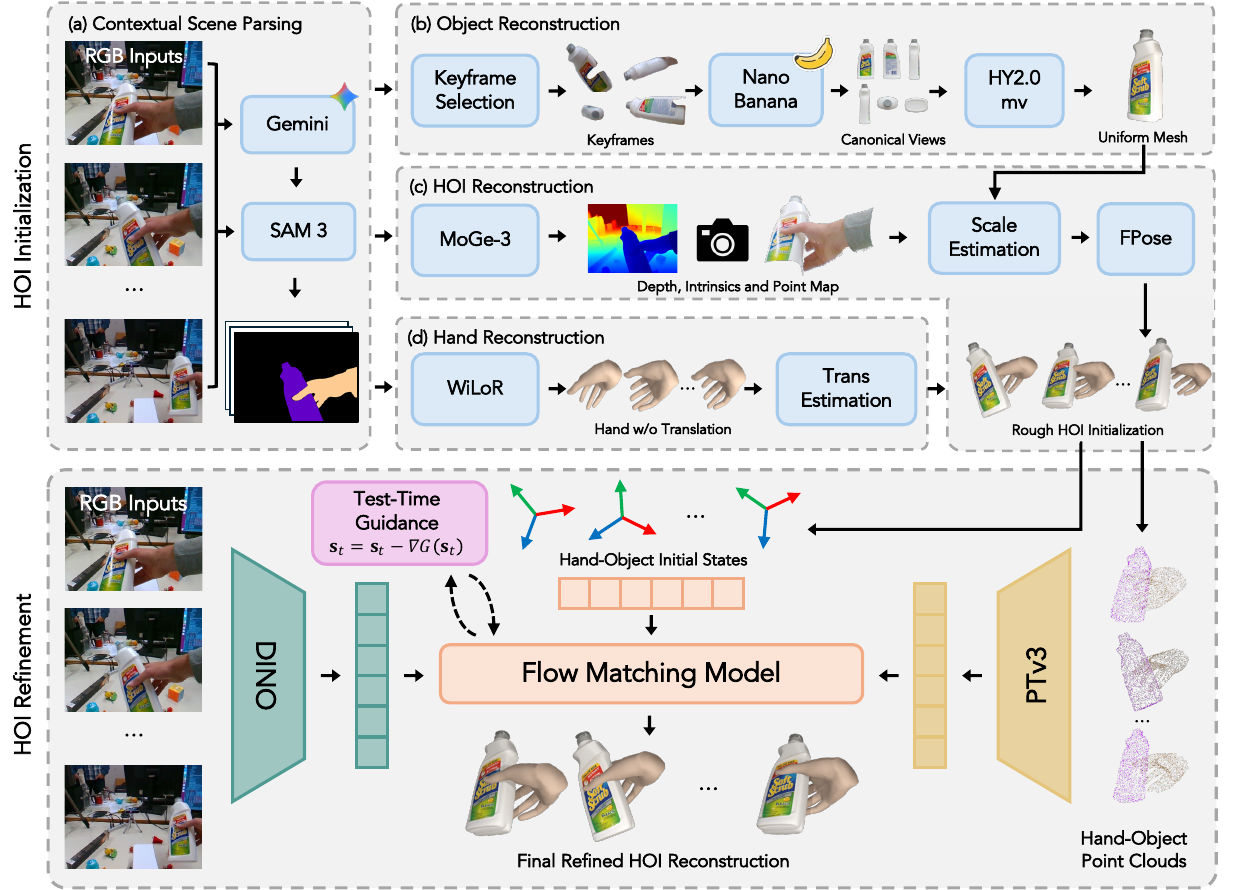}
    \vspace{-12pt}
    \caption{\small \textbf{\ourmethod overview.} Given a monocular video, we construct initial hand-object states by leveraging foundation models, including (a) contextual scene parsing, (b) object reconstruction, (d) hand reconstruction, and (c) depth alignment for recovering metric geometry, producing coarse HOI initialization. We then formulate HOI refinement as a conditional generative bridge matching problem that takes the HOI initialization together with RGB and 3D cues as input and outputs the refined HOI reconstruction. At inference time, test-time guidance adjusts the evolving states to better satisfy physical and image-space constraints.}
    \label{fig:framework}
\end{figure}

\textbf{Flow matching for HOI reconstruction.}\label{sec:flow_matching}
Flow matching~\citep{lipman2022flow, liu2022flow} formulates generative modeling as learning a continuous transport from a source distribution $p(\mathbf{x}_0)$ to a target data distribution $p(\mathbf{x}_1)$. Let $\mathbf{x}_0 \sim p(\mathbf{x}_0)$ and $\mathbf{x}_1 \sim p(\mathbf{x}_1)$. A stochastic interpolant defines intermediate states
$
\mathbf{x}_t=(1 - t)\mathbf{x}_0 + t\mathbf{x}_1 + \gamma(t)\boldsymbol{\epsilon},
$
where $t \in [0,1]$, $\boldsymbol{\epsilon} \sim \mathcal{N}(0, \mathbf{I})$, and $\gamma(t)$ is a noise schedule. Flow matching learns a velocity field $\mathbf{v}_\theta(\mathbf{x}, t)$ such that the trajectory follows $d\mathbf{x}(t) / dt = \mathbf{v}_\theta(\mathbf{x}(t), t)$,
transporting samples from $p(\mathbf{x}_0)$ to $p(\mathbf{x}_1)$. In practice, $\mathbf{v}_\theta$ can be a neural network (e.g., DiT~\citep{peebles2023scalable}) trained to match the conditional drift induced by the interpolate via the following objective:
\begin{align}
\mathcal{L}_\theta 
&= \mathbb{E}_{t,\mathbf{x}_0,\mathbf{x}_1,\boldsymbol{\epsilon}}
\left[
\left\|
\mathbf{v}_\theta(\mathbf{x}_t, t) - (\mathbf{x}_1 - \mathbf{x}_t)/(1 - t)
\right\|_2^2
\right]. \label{eq:flow_matching}
\end{align}
In the context of HOI, the target distribution $p(\mathbf{x}_1)$ corresponds to physically plausible hand-object interactions, which can be learned from large-scale HOI datasets~\citep{wang2024ho,yang2022oakink,swamy2023showme,hampali2020honnotate}.
While several recent works adopt generative models for HOI, they often start from random noise~\citep{ye2022s,ye2026whole}, depend on object templates~\citep{ye2024g}, or target the task of grasp synthesis~\citep{wu2026affordgrasp} instead of full 4D HOI reconstruction. 

\textbf{Bridge matching.}
A key limitation of vanilla flow matching is that it assumes a fixed Gaussian prior, whose stochastic nature can hinder physical accuracy in many deterministic tasks. To address this, the Bridge Matching~\citep{albergo2025stochastic} framework alleviates this by learning transport between arbitrary distributions $\pi_0$ and $\pi_1$. Given $\mathbf{x}_0 \sim \pi_0$ and $\mathbf{x}_1 \sim \pi_1$, the trajectory can be written as
\begin{align}
d\mathbf{x}_t = \mathbf{v}_\theta(\mathbf{x}_t, t)\,dt + \sigma(t)\,d\mathbf{W}_t,
\end{align}
where $\mathbf{v}_\theta$ denotes the learnable drift and $\sigma(t)$ controls the noise level. 
A common instantiation is the Brownian bridge with $\sigma(t)=\sigma\sqrt{t(1-t)}$. The velocity field is trained by matching the drift:
\begin{align}
\mathcal{L}_{\text{bridge}} = 
\mathbb{E}_{t, \mathbf{x}_0,\mathbf{x}_1}
\left[
\left\|
\mathbf{v}_\theta(\mathbf{x}_t, t) - (\mathbf{x}_1 - \mathbf{x}_t)/(1 - t)
\right\|_2^2
\right].\label{eq:bridge}
\end{align}
When $\sigma(t)=0$, Eq~(\ref{eq:bridge}) reduces to deterministic flow matching. By replacing the Gaussian prior with a more plausible initial distribution, it reduces the stochasticity inherent in noise-driven generation. A key remaining question therefore is the choice of the source distribution. In this work, we define it using foundation-model-derived hand and object states, which will be described in the next section.

\subsection{Hand and Object States Initialization} \label{ss:fm}
The goal is to leverage foundation models to obtain initial hand and object geometry and their per-frame interaction initialization $\mathbf{s}_0^{i} = [\mathbf{h}_0^{i}, \mathbf{o}_0^{i}]$, where $i$ indexes video frames. For brevity, we will omit the frame index $i$. These estimates provide coarse yet informative hand and object pose, which serve as the starting point of our conditional generative bridge. While foundation models has made significant progress,
they treat hand, object, and the context separately and do not consider their intricate interactions. To address this, we introduce the following methods for reconstructing hand, object, and their interaction. Please refer to Appendix for the detailed process. 

\textbf{Contextual scene parsing.}
We first perform holistic 3D scene understanding to reduce ambiguity in monocular reconstruction and generate outputs for object and hand states estimation. Specifically, for each frame, we obtain the object mask $\mathbf{M}_i$ via vision language grounding~\citep{comanici2025gemini} and segmentation models~\citep{carion2026sam}. We also estimate depth $\mathbf{D}_i$ and camera parameters using MoGe-3~\citep{kong2026moge}, which provide geometric cues for both hand and object reconstruction. These outputs establish spatial consistency across frames. Figure~\ref{fig:framework}-(a) illustrates the main idea. 

\textbf{Hand reconstruction and state estimation.}
For hand initialization, we employ WiLoR~\citep{potamias2025wilor} to estimate per-frame MANO~\citep{romero2022embodied} parameters and 2D keypoints, with hand bounding boxes extracted from masks. While WiLoR provides reliable hand articulation and rotation estimates, its translation prediction is limited by the weak-perspective camera assumption, which cannot accurately recover metric camera-space depth. To address this, we solve a lightweight closed-form least-squares to estimate the hand translation. Figure~\ref{fig:framework}-(d) shows the pipeline. 

\textbf{Object reconstruction and state estimation.}\label{sec:object}
We estimate a canonical object mesh $\mathbf{m}_o$ and temporally consistent per-frame poses $\{\mathbf{o}^i\}_{i=1}^N$. We sample sparse keyframes with diverse viewpoints and reconstruct the mesh in normalized space using pre-trained multi-view generative models~\citep{comanici2025gemini,zhao2025hunyuan3d}. Compared with single-view generation~\citep{chen2026sam}, this reduces ambiguity under occlusion and improves texture quality. We then align the object with the scene using depth cues and an enhanced FoundationPose~\citep{wen2024foundationpose}, as shown in Figure~\ref{fig:framework}-(b). The resulting states provide plausible structural alignment but remain susceptible to occlusion-induced instability and generative artifacts, motivating subsequent bridge matching refinement.

\subsection{4D HOI Flow Matching} \label{ss:gen_model}
After we obtained initialized hand-object states $\mathbf{s}_{0}$ from the input video sequence, we learn a conditional generative model that transports these states toward physically plausible interactions. We parameterize the conditional velocity field $\mathbf{v}_\theta(\mathbf{s}_t, t)$ using a neural network that takes as input the current state $\mathbf{s}_t$, flow time $t$, and multi-modal conditions derived from the input video.

\textbf{Conditional bridge matching.}\label{sec:model}
As discussed in Section~\ref{sec:flow_matching}, directly applying Eq.~(\ref{eq:flow_matching}) to HOI reconstruction is challenging due to the large solution space and ambiguity of monocular observations. To address this issue, we inject an initialization state $\mathbf{s}_{0}$ through the bridge matching formulation in Eq.~(\ref{eq:bridge}), which guides the transport process toward physically plausible hand-object configurations.

We further encode geometric structure by representing both the hand and object as point clouds. Specifically, we sample points from the object mesh $\mathbf{m}_o$ and use MANO vertices parameterized by shape coefficients $\bm{\beta}$ for the hand representation. These point clouds are processed using a Point Transformer V3 backbone~\citep{wu2024point} to capture spatial relationships between the hand and object.
To incorporate visual context, we extract image features from each frame using a pretrained DINOv2 encoder~\citep{oquab2023dinov2}.
For efficient 4D modeling, we aggregate multi-frame features and feed them into a DiT backbone for temporal reasoning and conditional flow prediction.
Detailed architecture is shown in Figure~\ref{fig:model}.
The bridge matching objective in Eq.~(\ref{eq:bridge})  can be written as:
\begin{align}
\mathcal{L}_{\text{hof}} = 
\mathbb{E}_{\mathbf{s}_0,\mathbf{s}_1,t}
\left[
\left\|
\mathbf{v}_\theta(\mathbf{s}_t, t, \mathbf{m}_o, \bm{\beta}, \mathcal{I}) - ({\mathbf{s}_1 - \mathbf{s}_t})/({1 - t})
\right\|_2^2
\right],
\label{eq:hof}
\end{align}
where $\mathbf{m}_o$ denotes the object mesh, $\bm{\beta}$ the hand shape, and $\mathcal{I}$ the sequence of input images. This objective learns to transport the initialized states $\mathbf{s}_0$ toward the target interaction distribution.

\textbf{Training objectives.}
We introduce three auxiliary losses to improve reconstruction quality: 3D position loss $\mathcal{L}_\text{3D}$, interaction loss $\mathcal{L}_{\text{inter}}$, and regularization loss $\mathcal{L}_\text{reg}$.

We apply regression-based~\citep{pavlakos2024reconstructing} 3D position losses $\mathcal{L}_\text{3D}$ to the terminal estimate $\mathbf{s}_1$, recovered from the generated trajectory since flow matching predicts velocities rather than states.
The 3D loss consists of three components: (1) a hand loss supervising the 3D MANO joints $\mathbf{J}_\text{3D}$, (2) an object loss supervising the transformed object point cloud $\mathbf{P}$, and (3) a relative loss supervising the spatial relationship between the hand and the object through their relative Euclidean distances:
\begin{align}
    \mathcal{L}_\text{3D}=\left\|\mathbf{J}_\text{3D}-\mathbf{J}^*_\text{3D}\right\|_1+\left\|\mathbf{P}-\mathbf{P}^*\right\|_1+\left\|\mathbf{D}(\mathbf{P},\mathbf{J}_\text{3D})-\mathbf{D}(\mathbf{P}^*,\mathbf{J}_\text{3D}^*)\right\|_1,
\end{align}
where $\mathbf{P}$ is the point cloud sampled from the transformed object mesh $\mathbf{m}_o$, $\mathbf{D}(\mathbf{P},\mathbf{J}_\text{3D})$ is the distance matrix for each transformed object point-hand joint pair. The asterisk denotes the ground truth.

To encourage physically plausible interactions, we further incorporate contact and penetration-based constraints~\citep{jiang2021hand}, denoted by $\mathcal{L}_{\text{inter}}$, between the hand mesh and object point cloud:
\begin{align}
    \mathcal{L}_{\text{inter}}=\sum\nolimits_{j\in\Omega}d(p_j,\mathcal{H})+\sum\nolimits_{j\in\mathcal{P}}d(p_j,\mathcal{H}),
\end{align}
where $d(p_j,\mathcal{H})$ is the distance between object point $p_j$ to MANO mesh surface $\mathcal{H}$, $\Omega$ is the contact set derived from ground-truth pose and $\mathcal{P}$ is the object point set that penetrates into the hand mesh. The first term encourages contact at ground-truth contact regions, while the second penalizes penetration.

In addition, we empirically observe that the hand estimates provided by the initialization are accurate in most cases. Therefore, the refinement process should make only subtle adjustments to improve physical plausibility, rather than performing unconstrained optimization that may introduce undesirable artifacts. To this end, we introduce a regularization term, $\mathcal{L}_{\text{reg}}$, on hand articulation related channels $\mathbf{v}_{\theta, \text{hand}}$ to preserve the initialized hand motion during refinement:
\begin{align}
    \mathcal{L}_{\text{reg}}=\left\|\mathbf{v}_{\theta, \text{hand}}(\mathbf{s}_t, t, \mathbf{m}_o, \bm{\beta}, \mathcal{I})\right\|_2^2.
\end{align}

The overall training objective combines the bridge matching loss with the auxiliary terms:
\begin{align}
    \mathcal{L}_\text{total} = \mathcal{L}_\text{hof} + \mathcal{L}_\text{3D} + \mathcal{L}_\text{inter} + \mathcal{L}_\text{reg}.
\end{align}

\textbf{Test-time guidance.} 
An additional benefit of our bridge matching formulation is that the continuous transport process provides a flexible interface for test-time control. During inference, task-specific constraints can be directly incorporated by steering intermediate states along the learned transport trajectory, without performing expensive per-sequence optimization. This allows us to efficiently impose additional geometric and interaction constraints, such as contact, temporal consistency, and image-space alignment, while preserving the interaction prior learned by the model.
Formally, we incorporate the following lightweight test-time guidance $G(\mathbf{s}_t)$ into the bridge matching process: 
\begin{align}
    \mathbf{s}_t = \mathbf{s}_t-\nabla_{\mathbf{s}_t} G(\mathbf{s}_t).
\end{align}
The guidance function $G$ combines three complementary objectives: silhouette guidance $G_{\text{sil}}$ for image alignment, interaction guidance $G_{\text{inter}}$ for physical correctness, and temporal smoothness guidance $G_{\text{temp}}$ for motion consistency.

The silhouette guidance $G_{\text{sil}}$ aligns the projected
object geometry with the observed segmentation. We sample object surface
points and project them onto the image plane to obtain $\mathcal{U}$. Let
$\mathcal{B}$ denote sampled boundary pixels of the visible object mask, and
let $\mathcal{M}$ be the union of the object mask and the dilated hand mask.
The guidance is defined as
\begin{align}
G_{\text{sil}}
=
\frac{1}{2|\mathcal{U}|}
\sum_{\mathbf{u}\in\mathcal{U}}
\rho\!\left(
D_{\mathcal{M}}(\mathbf{u})
+
D_{\Psi}(\mathbf{u})
\right)
+
\frac{1}{2|\mathcal{B}|}
\sum_{\mathbf{b}\in\mathcal{B}}
\rho\!\left(
D_{\mathcal{U}}(\mathbf{b})
\right),
\end{align}
where $D_{\mathcal{M}}(\mathbf{u})$ is the distance from $\mathbf{u}$ to
the admissible region $\mathcal{M}$, $D_{\Psi}(\mathbf{u})$ 
projections outside the image, and $D_{\mathcal{U}}(\mathbf{b})$ is the
distance from $\mathbf{b}$ to its nearest projected point in $\mathcal{U}$.
Here, $\rho$ denotes the smooth L1 penalty. The first term penalizes projections outside the object or hand-occluded regions, while the second term encourages coverage of the visible object boundary.

Our interaction guidance $G_{\text{inter}}$ follows the interaction loss $\mathcal{L}_{\text{inter}}$ used during training, encouraging contact while penalizing interpenetration. Since ground-truth contact annotations are unavailable at test time, we use the learned contact prediction model HACO~\citep{jung2026learning} to infer the contact index set $\Omega$. This allows guidance to adapt to diverse hand-object interaction patterns without relying on manually predefined fixed contact regions~\citep{fan2024hold, hasson2019learning}.

To further reduce temporal jitter, we introduce temporal smoothness guidance $G_{\text{temp}}$ that penalizes the velocities and accelerations of the hand and object:
\begin{align}
G_{\text{temp}}=\|\nabla \mathbf{s}^i\|_2^2 + \|\nabla^2 \mathbf{s}^i\|_2^2,
\end{align}
where $\nabla$ and $\nabla^2$ denote first- and second-order temporal finite differences, respectively. Translational increments are computed as differences between consecutive positions, while rotational increments are represented by the axis-angle vectors of the corresponding relative rotations. Second-order differences are computed from consecutive increments.

We apply guidance only during the later stages of flow-matching integration, when the predicted state is already close to a plausible interaction and requires only minor corrections. By concentrating refinement on these later steps, the guidance makes targeted adjustments that improve image alignment, interaction validity, and temporal consistency while introducing limited computational overhead.

\subsection{Data Preparation and Model Training} \label{ss:training}
We train our model on multiple large-scale HOI datasets, including HO-Cap~\citep{wang2024ho}, OakInk~\citep{yang2022oakink}, and SHOWMe~\citep{swamy2023showme}, which covering diverse interactions and object categories. Training excludes all evaluation benchmarks, and we use identical model weights across test datasets to assess cross-dataset generalization.

To efficiently leverage the available data and prevent overfitting to specific dataset pattern, we construct training pairs by perturbing the ground-truth HOI states $\mathbf{s}_1$ with a randomly sampled noise (see Appendix for details). 
Unlike conventional generative modeling that transports samples from pure Gaussian noise, our formulation learns a conditional transport process between imperfect yet structured initialization states and physically plausible HOI states. This matches the practical inference setting, where modern foundation models already provide strong but imperfect estimates.
 
\begingroup
\setlength{\intextsep}{6pt plus 2pt minus 2pt}
\section{Experiments}

\subsection{Setup}
\mypara{Datasets.}
We evaluate our approach on two standard benchmarks: HO3D-v3~\citep{hampali2020honnotate} and DexYCB~\citep{chao2021dexycb}. HO3D-v3 features complex interactions resulting in severe occlusions, and DexYCB emphasizes tabletop grasping across diverse subjects. 
Following AGILE's evaluation protocol~\citep{shi2026agile}, we use 18 HO3D-v3 and 20 DexYCB sequences; we refer readers to AGILE for sequence selection and evaluation details.

\mypara{Baselines.}
We compare with state-of-the-art monocular HOI reconstruction methods: HOLD~\citep{fan2024hold} jointly optimizes implicit hand-object fields; MagicHOI~\citep{wang2025magichoi} improves object geometry with novel-view synthesis priors. AGILE~\citep{shi2026agile} combines agentic initialization with heuristic optimization, and CHOIR~\citep{xu2026choir} integrates foundation-model initialization, learned relative-depth correction, and contact-aware joint optimization.

\mypara{Metrics.}
We evaluate five complementary aspects. \emph{Hand accuracy}: mean per-joint position error (MPJPE) [mm] for local articulation. \emph{Object accuracy}: bidirectional Chamfer distance (CD) [cm$^2$] and F-scores at 5/10\,mm (F5/F10) [\%] after ICP alignment of scale, rotation, and translation. \emph{Interaction consistency}: hand-relative Chamfer distance (CD$_h$) [cm$^2$]~\citep{fan2024hold}, computed in the hand-root frame; interpenetration volume (IV) [cm$^3$], measuring hand-object overlap; and contact accuracy (CA) [\%]~\citep{jiang2021hand}, the fraction of frames whose contact states match the ground truth. \emph{Temporal stability}: hand-joint trajectory jerk magnitude (JM) [m/s$^3$], with lower values indicating smoother motion. \emph{Robustness}: per-sequence success rate (SR) [\%]~\citep{shi2026agile}, requiring both hand and object reconstruction to complete without failure.

\subsection{4D HOI Reconstruction Comparison}
We compare \ourmethod against state-of-the-art methods on HO3D-v3 (Table~\ref{tab:ho3d}) and DexYCB (Table~\ref{tab:dexycb}), and show qualitative results in Figure~\ref{fig:qual} and the supplementary video.

\begin{table}[t]
\small
\caption{\small Results on HO3D-v3. Our method shows strong improvements on \emph{interaction} metrics, which are often underreported in previous methods~\citep{wang2025magichoi,shi2026agile}. Hand and object accuracy are determined by foundation models and our method performs on par with AGILE (code not available). 
}
\vspace{-1mm}
\label{tab:ho3d}
\centering
\small
    \begin{tabular}{lccccccccc}
    \toprule
    \multirow{2.5}{*}{Method} & \multicolumn{4}{c}{Interaction} & Hand & \multicolumn{3}{c}{Object} & Robustness \\
    \cmidrule(r){2-5} \cmidrule(r){6-6} \cmidrule(r){7-9} \cmidrule(r){10-10}
    & CD$_h$ $\downarrow$ & IV $\downarrow$ & CA $\uparrow$ & JM $\downarrow$ & MPJPE $\downarrow$ & CD $\downarrow$ & F5 $\uparrow$ & F10 $\uparrow$ & SR $\uparrow$ \\
    \midrule
    HOLD     & \cellcolor{third}18.66 & \cellcolor{third}18.76 & \cellcolor{third}89.50 & \cellcolor{third}10.43 & 22.09 & 1.11 & \cellcolor{third}81.75 & \cellcolor{third}92.42 & \cellcolor{best}\textbf{100.0} \\
    MagicHOI & 21.81 & \cellcolor{second}17.41 & 83.00 & 21.66 & \cellcolor{third}7.38  & 0.90 & 76.74 & 91.59 & \cellcolor{third}83.3 \\
    AGILE   & \cellcolor{second}{15.81} & --     & -- & --     & \cellcolor{best}\textbf{3.92}  & \cellcolor{best}\textbf{0.27} & \cellcolor{second}{86.63} & \cellcolor{best}\textbf{97.77} & \cellcolor{best}\textbf{100.0}\\
    CHOIR & 42.37 & 22.90 & \cellcolor{second}90.99 & \cellcolor{best}\textbf{0.92} & 21.12  & \cellcolor{third}0.84 & 63.87 & 90.29 & \cellcolor{second}94.4 \\
    \textbf{\ourmethod (ours)}     & \cellcolor{best}\textbf{5.98} & \cellcolor{best}\textbf{13.44} & \cellcolor{best}\textbf{97.27} & \cellcolor{second}{1.61} & \cellcolor{second}{4.40}  & \cellcolor{second}{0.28} & \cellcolor{best}\textbf{89.11} & \cellcolor{second}{97.13} & \cellcolor{best}\textbf{100.0} \\
    \bottomrule
    \end{tabular}
\end{table}
\begin{table}[!htb]
\begin{minipage}[t]{0.64\linewidth}
\vspace{0pt}
\small
\setlength{\tabcolsep}{3pt}
\caption{\small Results on DexYCB~\citep{chao2021dexycb} dataset.}
\vspace{-3mm}
\label{tab:dexycb}
\centering  
    \resizebox{\linewidth}{!}{%
    \begin{tabular}{lcccccc}
    \toprule
    \multirow{2.5}{*}{Method}    & Hand  & \multicolumn{3}{c}{Object} & Interaction & Robustness \\ 
    \cmidrule(r){2-2} \cmidrule(r){3-5} \cmidrule(r){6-6} \cmidrule(r){7-7}
    & MPJPE $\downarrow$ & CD  $\downarrow$ & F5 $\uparrow$ & F10 $\uparrow$ & CD$_h$ $\downarrow$ & SR $\uparrow$ \\ 
    \midrule
    HOLD & 30.86    & 19.30  & 33.20 & 54.94  & 170.90  & \cellcolor{second}{45.0}     \\ 
    MagicHOI    &\cellcolor{third}21.20    & 2.05  & 45.67 & 67.14  & 661.90  & \cellcolor{third}25.0      \\ 
    AGILE    & \cellcolor{second}{19.06}    & \cellcolor{second}{0.52}  & \cellcolor{second}{83.21} & \cellcolor{second}{95.43}  & \cellcolor{second}{94.60}   & \cellcolor{best}\textbf{100.0}      \\ 
    CHOIR    &  27.24   & \cellcolor{third}0.84  & \cellcolor{third}79.46 & \cellcolor{third}93.09  &  \cellcolor{third}134.74  &  \cellcolor{best}\textbf{100.0}    \\
    \textbf{\ourmethod (ours)}     & \cellcolor{best}\textbf{6.84}   & \cellcolor{best}\textbf{0.45}  & \cellcolor{best}\textbf{83.62} & \cellcolor{best}\textbf{95.81}  & \cellcolor{best}\textbf{48.85}   & \cellcolor{best}\textbf{100.0}      \\ 
    \bottomrule
    \end{tabular}%
    }
    \vspace{-2mm}

\end{minipage}
\hfill
\begin{minipage}[t]{0.34\linewidth}
\vspace{0pt}
    \centering
    \small
    \renewcommand{\arraystretch}{1.48}
    \setlength{\tabcolsep}{2pt}
    \caption{\small {Computational cost.}} 
    \vspace{-3mm}
    \label{tab:time}
    \resizebox{\linewidth}{!}{%
    \begin{tabular}{@{}lccc@{}}
        \toprule
        Method & Preproc. & Optim. & Total \\
        \midrule
        \ourmethod{} & $\sim$23 min & $\sim$1 min
                    & $\sim$24 min \\
        \midrule
        HOLD        & $\sim$65 min & $\sim$24 h
                    & $\sim$25.1 h \\
        MagicHOI    & $\sim$25 min & $\sim$2 h
                    & $\sim$2.4 h \\
        AGILE       & 21--29 min   & $\sim$1.75 h
                    & 2.1--2.2 h \\
        CHOIR       & $\sim$11min  & $\sim$19min 
                    & $\sim$30min \\
        \bottomrule
    \end{tabular}%
    }
    \vspace{-3mm}

\end{minipage}
\end{table}

\textbf{Results on HO3D-v3.}
\ourmethod attains the lowest hand-to-object Chamfer distance and interpenetration volume among all methods, improving over the next best baseline by 62.2\% in CD$_h$. The jerk magnitude is reduced to a level $6.4\times$ lower than HOLD~\citep{fan2024hold} and $13.4\times$ lower than MagicHOI~\citep{wang2025magichoi}, confirming that our learned interaction prior produces temporally stable trajectories without explicit smoothness regularization. Our method further achieves the highest contact ratio among reporting methods and the strongest F5 among all baselines.

\textbf{Results on DexYCB.}
DexYCB contains diverse subjects performing tabletop grasps, where prior baselines suffer from large scale and translation errors. 
\ourmethod yields a particularly pronounced margin across all the metrics comparing with other baselines.

\begin{figure}[t]
    \centering
    \small
    \includegraphics[width=\textwidth]{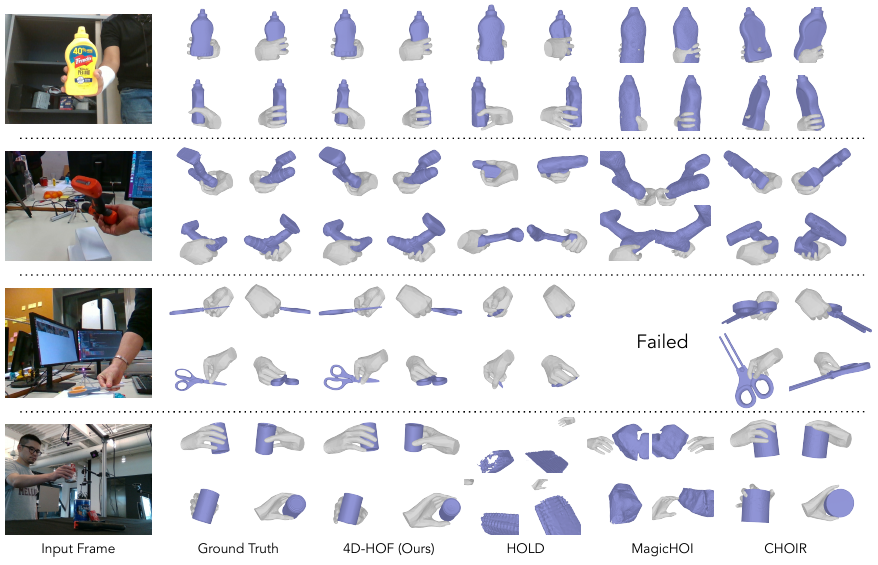}
    \vspace{-20pt}
    \caption{\small {Qualitative results on HO3D-v3~\citep{hampali2020honnotate} and DexYCB~\citep{chao2021dexycb}.} \ourmethod consistently outperforms HOLD~\citep{fan2024hold} and MagicHOI~\citep{wang2025magichoi} in preserving accurate geometry and hand-object interactions. Please see our supplementary video for detailed comparisons.}
    \vspace{-3mm}
    \label{fig:qual}
\end{figure}

\textbf{Computational cost.}
We compare the inference time in Table~\ref{tab:time}. Existing methods rely on expensive test-time optimization and typically require hours to process a 200-frame sequence. In contrast, \ourmethod performs reconstruction through a feed-forward generative transport process, reducing the refinement stage to approximately one minute while maintaining competitive reconstruction quality.

\subsection{Ablation Study and Analysis}
We conduct ablation studies on HO3D-v3~\citep{hampali2020honnotate} to validate the design choices. 


\begin{table*}[t]
\centering

\begin{minipage}[t]{0.49\textwidth}
\centering
\renewcommand{\arraystretch}{1.4}
\caption{\small {Ablation of vision foundation models.}}
\label{tab:ablation}
\vspace{-2mm}
\resizebox{\linewidth}{!}{%
\begin{tabular}{lccccc}
\toprule
\multirow{2.5}{*}{Method Variants}
    & \multicolumn{4}{c}{Interaction} & Hand \\
\cmidrule(lr){2-5} \cmidrule(lr){6-6}
    & CD$_h$ $\downarrow$ & IV $\downarrow$
    & CA $\uparrow$ & JM $\downarrow$ & MPJPE $\downarrow$ \\
\midrule
\ourmethod{} (full model)
    & 5.98 & 13.44 & 97.27 & 1.61 & 4.40 \\
\midrule
(a) SAM3D object recon.
    & 262.19 & 9.87 & 64.40 & 1.75 & 6.09 \\
(b) FPose default tracking
    & 15646.78 & 9.39 & 75.74 & 2.07 & 5.00 \\
(c) VFM only
    & 46.58 & 11.06 & 78.07 & 3.95 & 4.03 \\
(d) VFM + guidance
    & 41.75 & 11.89 & 81.53 & 3.71 & 4.03 \\
\bottomrule
\end{tabular}%
}
\end{minipage}
\hfill
\begin{minipage}[t]{0.49\textwidth}
\centering
\caption{\small {Ablation of the bridge matching model.}}
\label{tab:bridge_ablation}
\vspace{-2mm}
\resizebox{\linewidth}{!}{%
\begin{tabular}{lccccc}
\toprule
\multirow{2.5}{*}{Method Variants}
    & \multicolumn{4}{c}{Interaction} & Hand \\
\cmidrule(lr){2-5} \cmidrule(lr){6-6}
    & CD$_h$ $\downarrow$ & IV $\downarrow$
    & CA $\uparrow$ & JM $\downarrow$ & MPJPE $\downarrow$ \\
\midrule
\ourmethod{} (full model)
    & 5.98 & 13.44 & 97.27 & 1.61 & 4.40 \\
\midrule
(a) w/o temporal design
    & 18.93 & 14.23 & 95.64 & 4.26 & 4.15 \\
(b) w/o explicit encoding
    & 11.92 & 14.76 & 96.41 & 1.62 & 4.40 \\
(c) w/o object optimization
    & 90.49 & 9.63 & 62.23 & 2.31 & 5.06 \\
(d) w/o hand optimization
    & 12.34 & 13.05 & 96.02 & 3.95 & 4.03 \\
(e) w/o guidance
    & 8.03 & 13.39 & 96.62 & 1.71 & 4.37 \\
\bottomrule
\end{tabular}%
}
\end{minipage}

\vspace{-3mm}
\end{table*}

\textbf{How to use vision foundation models?}
We first study different strategies for constructing the initial hand-object states (Table~\ref{tab:ablation}). (a) Replacing our multi-view object generation with SAM3D~\citep{chen2026sam} significantly degrades interaction quality, showing the importance of temporally consistent object reconstruction. (b) Using FoundationPose~\citep{wen2024foundationpose} in pure tracking mode also fails under severe occlusions, where a single mistracked frame can corrupt the entire sequence. (c) Directly using vision foundation model outputs without our generative bridge yields poor alignment and contact. (d) Adding guidance (but without bridge matching model) offers only modest improvements, suggesting the need for a learned interaction prior to refine independently estimated HOI states. 

\textbf{How to design the HOI flow matching model?}
We next analyze the design of the conditional bridge (Table~\ref{tab:bridge_ablation}). (a): Removing temporal modeling produces unstable 4D trajectories, suggesting that independent frame-wise refinement cannot capture interaction dynamics. (b): Removing the explicit hand-object structural encoding consistently degrades interaction metrics, highlighting the importance of explicitly modeling relative hand-object geometry. (c) and (d): Optimizing only the hand or only the object also leads to worse interaction quality, indicating that hand and object states must be refined jointly to resolve their mutual ambiguity. (e): Removing test-time guidance degrades contact and interaction quality, showing that guidance provides effective test-time control toward geometrically and physically consistent HOI states.
Notably, these variants leave the object geometry unchanged, while contact quality and physical interaction degrade significantly.

\begin{table*}[t]
\centering
\begin{minipage}[t]{0.49\textwidth}
\centering
\caption{\small {Ablation studies on loss functions.}}
\label{tab:loss_ablation}
\vspace{-2mm}
\resizebox{\linewidth}{!}{%
\begin{tabular}{lccccc}
\toprule
\multirow{2.5}{*}{$\mathcal{L}$ Variants}
    & \multicolumn{4}{c}{Interaction} & Hand \\
\cmidrule(lr){2-5} \cmidrule(lr){6-6}
    & CD$_h$ $\downarrow$ & IV $\downarrow$
    & CA $\uparrow$ & JM $\downarrow$ & MPJPE $\downarrow$ \\
\midrule
\ourmethod
    & 5.98 & 13.44 & 97.27 & 1.61 & 4.40 \\
\midrule
w/o $\mathcal{L}_{\text{3D}}$
    & 7.68 & 15.94 & 96.27 & 2.93 & 3.98 \\
w/o $\mathcal{L}_{\text{inter}}$
    & 6.02 & 13.78 & 97.00 & 1.63 & 4.60 \\
w/o $\mathcal{L}_{\text{reg}}$
    & 6.71 & 13.78 & 97.15  & 1.10 & 5.64\\
\bottomrule
\end{tabular}%
}
\end{minipage}
\hfill
\begin{minipage}[t]{0.49\textwidth}
\centering
\caption{\small {Ablation studies on guidance terms.}}
\label{tab:guidance_ablation}
\vspace{-2mm}
\resizebox{\linewidth}{!}{%
\begin{tabular}{lccccc}
\toprule
\multirow{2.5}{*}{$G$ Variants}
    & \multicolumn{4}{c}{Interaction} & Hand \\
\cmidrule(lr){2-5} \cmidrule(lr){6-6}
    & CD$_h$ $\downarrow$ & IV $\downarrow$
    & CA $\uparrow$ & JM $\downarrow$ & MPJPE $\downarrow$ \\
\midrule
\ourmethod
    & 5.98 & 13.44 & 97.27 & 1.61 & 4.40 \\
\midrule
w/o $G_{\text{sil}}$
    & 7.51 & 13.85 & 97.06 & 1.72 & 4.75 \\
w/o $G_{\text{inter}}$
    & 7.81 & 16.00 & 95.07 & 1.63 & 4.74 \\
w/o $G_{\text{temp}}$
    & 6.35 & 14.53 & 97.29 & 1.91 & 4.74 \\
\bottomrule
\end{tabular}%
}
\end{minipage}

\vspace{-3mm}
\end{table*}

\textbf{How to supervise \ourmethod training?}
We ablate each auxiliary losses in Table~\ref{tab:loss_ablation}. Removing $\mathcal{L}_{\text{3D}}$ causes severe degradation across most metrics, showing that explicit 3D supervision is critical for HOI reconstruction. Removing $\mathcal{L}_\text{inter}$ reduces contact quality and hand pose accuracy, indicating that the contact loss encourages meaningful physical interaction. Removing $\mathcal{L}_\text{reg}$ leads to the largest drop in hand pose, suggesting that regularization helps preserve the hand state from foundation models. 

\textbf{How to guide \ourmethod inference?}
We ablate each test-time guidance term in Table~\ref{tab:guidance_ablation}. Removing $G_{\text{sil}}$ degrades all metrics, suggesting that image-space alignment helps refine the reconstructed interaction. Removing $G_{\text{inter}}$ leads to the largest increase in interpenetration and the largest drop in contact accuracy, demonstrating its importance for plausible interactions. Removing $G_{\text{temp}}$ causes the largest increase in motion jitter, confirming that temporal guidance improves motion consistency.
\endgroup
\section{Conclusion and Future Works}
We propose \ourmethod, a novel feed-forward framework for 4D hand-object interaction reconstruction from monocular video. Our key innovation is to replace noise-driven generation with initialization derived from foundation models, and to learn a conditional generative bridge that transports these initial states toward physically plausible interactions. Our formulation enables joint refinement of hand, object, and their interactions in a feed-forward manner. Experiments show that our approach achieves strong performance in interaction quality while maintaining competitive hand and object accuracy, with significantly improved inference efficiency.
One limitation, shared by many foundation model-based methods, is its sensitivity to initialization failures (Figure~\ref{fig:failure}). Our current formulation focuses on single-hand interactions with rigid objects. Extending \ourmethod to more complex settings, such as bimanual interactions and articulated or deformable objects, is an important future direction.

\bibliography{iclr2027_conference}
\bibliographystyle{iclr2027_conference}

\newpage
\appendix

\DoToC

\newpage
\setcounter{figure}{0}
\setcounter{table}{0}
\setcounter{equation}{0}
\renewcommand\thesection{\Alph{section}}
\renewcommand{\thetable}{S\arabic{table}}  
\renewcommand{\thefigure}{S\arabic{figure}}
\renewcommand{\theequation}{S\arabic{equation}}

\section{Implementation Details}
In this section, we describe additional details of our method. 
We first present the hand and object state initialization pipeline based on various foundation models in Section \ref{ss:app_init}, followed by details of the flow-matching model in Section \ref{ss:app_gen_model}. 

\subsection{Hand and Object State Initialization}
\label{ss:app_init}

\subsubsection{Contextual scene parsing.}
Given an input video $\mathcal{I}$, we first employ a VLM~\citep{comanici2025gemini} to understand the interaction and identify the object manipulated by the human hand. Since processing all video frames is computationally prohibitive, we uniformly sample five frames from the video and prompt the VLM to describe the target object. The resulting object description, together with the prompt “hand,” is then provided to SAM 3~\citep{carion2026sam} for segmentation. 

\subsubsection{Hand reconstruction and state estimation.}
For hand estimation, we use WiLoR~\citep{potamias2025wilor} to predict the MANO~\citep{romero2022embodied} parameters and 2D keypoints. We then recover the hand translation in camera coordinates using the camera intrinsics from MoGe-3~\citep{kong2026moge, valassakis2024handdgp}. Specifically, for each joint, the perspective projection yields the 2D position $(u,v)$ of each 3D joint following equations: 
\begin{align}
u=f_x\frac{\mathbf{J}_x+\mathbf{t}_x}{\mathbf{J}_z+\mathbf{t}_z}+c_x, \\
v=f_y\frac{\mathbf{J}_y+\mathbf{t}_y}{\mathbf{J}_z+\mathbf{t}_z}+c_y, 
\end{align}
where $(\mathbf{J}_x, \mathbf{J}_y, \mathbf{J}_z)$ denotes the position of 3D joint in the canonical hand space, $(\mathbf{t}_x, \mathbf{t}_y, \mathbf{t}_z)$ represents the global hand translation, and $f_x, f_y, c_x, c_y$ are the camera intrinsic parameters. By aggregating constraints from all joints, we construct a linear system and solve it using least squares. 

\subsubsection{Object reconstruction and state estimation.}
In order to automatically select a compact yet informative set of observations for the object from the video, we propose a visual feature–based frame selection strategy. Specifically, we extract image features using a pre-trained DINOv2~\citep{oquab2023dinov2} image encoder. Since DINO only yields patch-level embeddings, we also apply AnyUp~\citep{wimmer2026anyup} to enhance their spatial resolution. We then pool the upsampled features within the object mask region to obtain an object descriptor for each frame. 

After having the object descriptors, the next step is picking the frames that are visually different. The frame with the largest object mask is first chosen as the anchor, and Farthest Point Sampling (FPS) is performed in the descriptor space to select a visual diverse set of frames. These selected frames are used as context prompts for canonical view generation, and the synthesized views are subsequently fed into a multi-view-to-3D model~\citep{zhao2025hunyuan3d} to recover the object mesh. Although the object geometry is generally reliable, textures produced by generative models are often unsatisfactory, preserving coarse color spots but failing on fine-grained patterns and text. To address this limitation, we construct two types of texture maps: one obtained directly from the generative model, which is suitable for unstructured objects with simple appearance, and another derived by projecting the canonical views~\citep{huang2025mv}, which better preserves detailed patterns for structured objects. We render the object under multiple views using both texture types and employ a VLM to automatically select the optimal variant. 

To better leverage FoundationPose~\citep{wen2024foundationpose} for object pose estimation, we initialize the pose on the frame with the largest object mask and propagate it bidirectionally throughout the sequence. Instead of relying on the default tracking mode of FoundationPose, in which a single occluded frame may lead to catastrophic error accumulation, we perform independent per-frame registration while explicitly rejecting pose hypotheses that deviate significantly from the previous estimate. Specifically, we use the geodesic distance between adjacent frame rotations as the filtering criterion and set the threshold to 60$^{\circ}$. 

We further employ silhouette-based filtering to further enforce visual consistency. For each pose hypothesis, we render an object mask $\tilde{\mathbf{M}^o}$ and compute its intersection-over-union (IoU) with the segmentation mask $\mathbf{M}^o$. To reduce the influence of hand occlusion, we exclude the hand mask $\mathbf{M}^h$ region before computing the IoU: 
\begin{align}
    \text{IoU}=\frac{(\tilde{\mathbf{M}}^o \cap \sim \mathbf{M}^h ) \cap (\tilde{\mathbf{M}}^o \cap \sim \mathbf{M}^h )}{(\tilde{\mathbf{M}}^o \cap \sim \mathbf{M}^h ) \cup (\tilde{\mathbf{M}}^o \cap \sim \mathbf{M}^h )}.
\end{align}
We retain only the top 30 pose hypotheses with the highest IoU scores. To further improve robustness, we enable both RGB and RGB-D modes in the pose refinement and scoring stages of FoundationPose, and compute the final pose score as the average of the two modalities. 

\begin{figure}[t]
    \centering
    \small
    \includegraphics[width=0.6\linewidth]{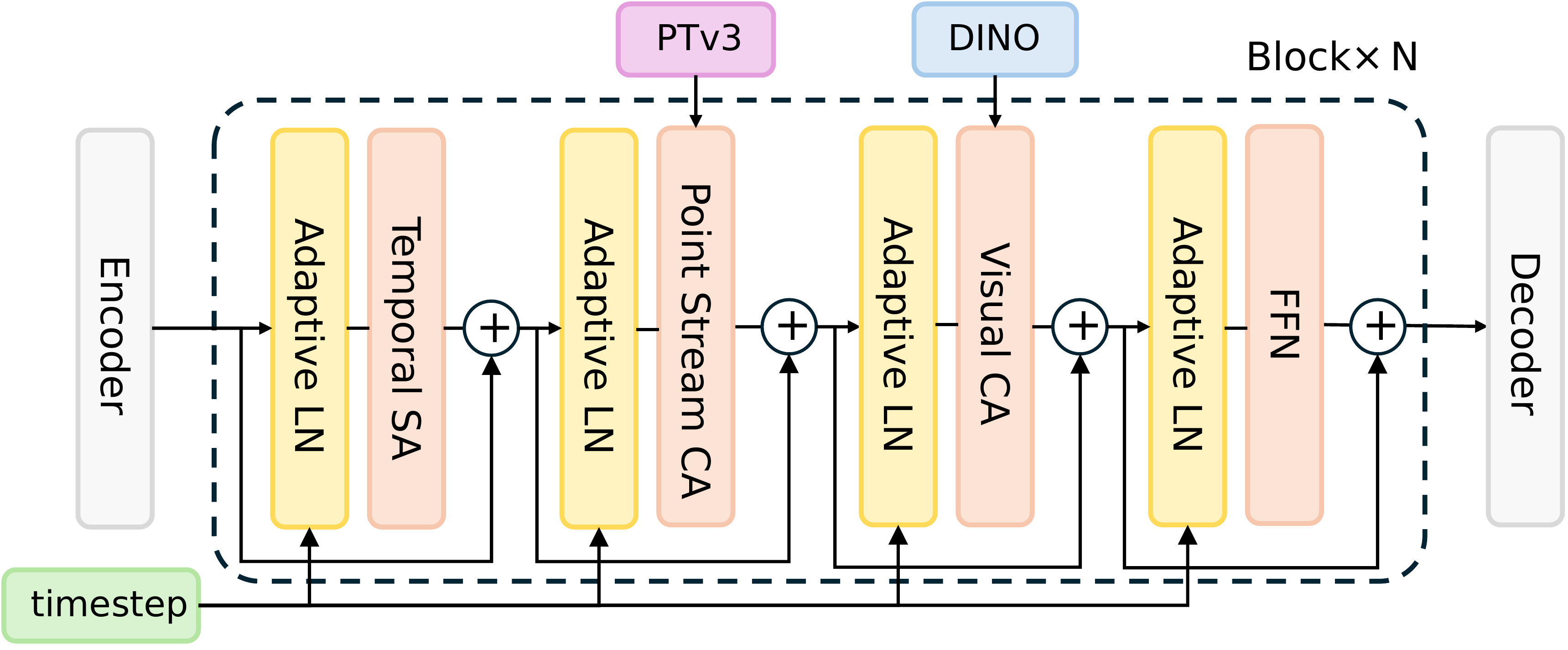}
    \caption{\small Model architecture. Our model adopts a DiT-style architecture for conditional flow matching. The latent hand-object state is processed through temporal self-attention, point-stream cross-attention using Point Transformer V3 features, and visual cross-attention using DINO features. Adaptive layer normalization conditioned on the timestep is applied throughout the network to modulate the generative transport process.}
    \label{fig:model}
\end{figure}

\subsection{4D HOI Flow Matching} \label{ss:app_gen_model}
\subsubsection{Model architecture}
Our flow matching model transports the noisy hand and object pose from the foundation model-based initialization to a plausible interaction pose manifold. Since we aim to reconstruct the 4D hand object interaction, a natural design is directly handling the sequence data, which benefits from the temporal information interaction. Specifically, our model takes a hand and object pose sequence as input. In each frame, the hand pose and object pose are concatenated together and tokenized by a linear layer. This pose token sequence is then passed thought several stacked Transformer blocks, after that the hand and object pose in each frame are decoded from the corresponding pose token. 

There are three attention operations in each Transformer block. Our block starts with self-attention operation along the temporal dimension with RoPE~\citep{su2024roformer} for temporal embedding. This role of this layer is to utilize the temporal corrections in the sequence. In a motion sequence, the pose divergence usually small for nearby frames and large for far away frames, besides the motion evolution in the sequence should be smooth. By learning large scale data, this layer can make the pose sequences coherent and smooth. The second layer is a cross-attention layer between the 3D HOI point cloud sequence and pose token. The motivation of this operation is to inject explicit 3D spatial information to the model, since it’s impossible to understand the hand object interaction solely from a set of bare rotations and translation. To prepare 3D cues, we use the pose results from the initialization stage to transform the MANO hand and object, then use a Point Transformer V3~\citep{wu2024point} to encode them. In each frame, we use MANO vertices as hand point cloud and sample points on the object surface as the object point cloud. We treat the two components as a single point cloud and add a class label to each point. After feature extraction, a max pooling operation is applied to construct the final HOI point cloud token sequence. By this cross attention, the model can obtain explicit 3D information which boosts joint optimization. The third one is another cross attention with the 2D image features, since we hope the reconstruction results respect the observation. For each video frame, we use a frozen DINOv2-B to extract visual features and use Plucker ray to add camera information to each patch embedding. To balance the computational cost, we only apply cross attention within each frame. A standard feed-forward layer is added at the end of block, and adaptive layer norm is used before each attention and FFN later to inject flow-step information. 

\subsubsection{Training data curation}
We initially explored training with initializations produced by our foundation model pipeline, but observed two limitations. First, object symmetries introduce ambiguities in the estimated rotations, complicating the construction of interpolation paths for bridge matching and subsequent training. Second, initialization errors exhibit dataset-specific biases, limiting the effectiveness of these initializations as a training source. We therefore adopt a dataset-agnostic perturbation strategy that synthetically corrupts ground-truth HOI states to emulate imperfect initializations.

We introduce synthetic noise to the ground-truth HOI states to emulate the imperfect initializations generated by the foundation models. To perturb the object pose, we sample a large offset (3.14 rad, 0.2 m) across 20\% of the frames to simulate catastrophic pose estimation failures, while applying a smaller offset (0.1 rad, 0.02 m) to the remaining 80\%. For the hand parameters, the translation vector is perturbed using noise sampled from a uniform spatial distribution (0.02, 0.02, 0.1). Meanwhile, the global orientation and the local pose of the hand are perturbed by adding Gaussian noise with a standard deviation of 0.1.

\subsubsection{Training details}
We optimize our model using AdamW with a learning rate of $0.0001$ and a weight decay of $0.0001$, regulated by a cosine learning rate scheduler with a warmup phase. The training process spans 20 epochs with a batch size of 16. During feature extraction, input images are resized to 224$\times$224 before being encoded by DINOv2. To establish an explicit 3D encoding, we uniformly sample 4096 points from the object mesh surface. Finally, the temporal sequence length is set to 16 frames during training and extended to 128 frames during inference. The entire training process takes approximately 30 hours on a single NVIDIA H100 GPU.

\begin{figure}[h]
    \centering
    \small
    \includegraphics[width=\textwidth]{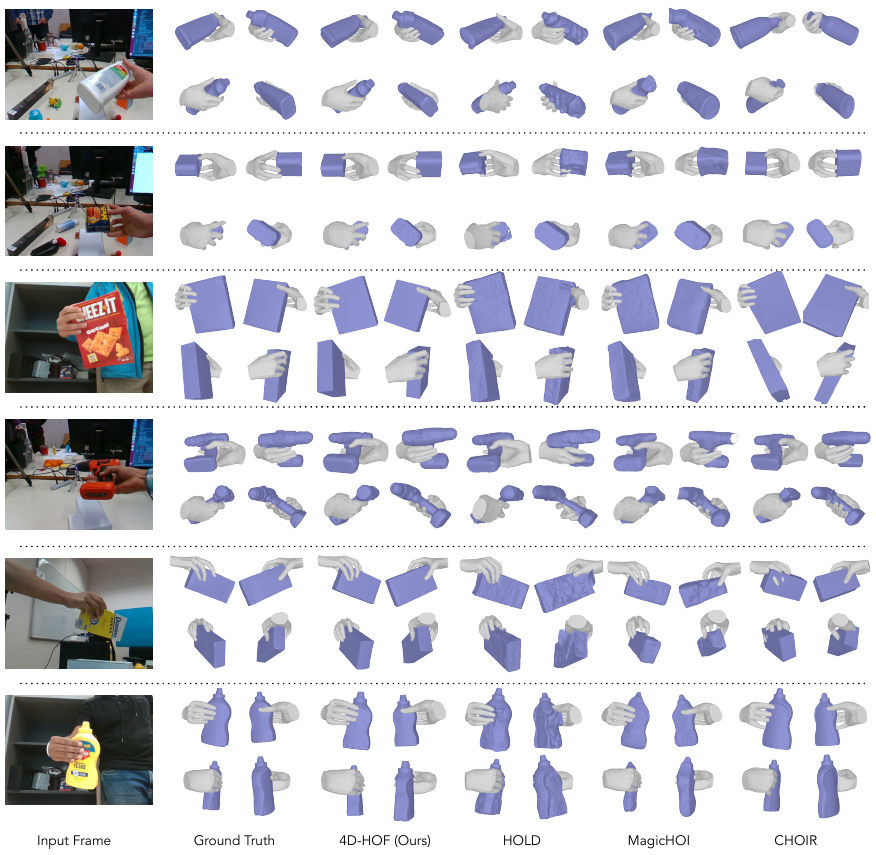}
    \vspace{-5mm}
    \caption{\small Qualitative results on HO3D-v3~\citep{hampali2020honnotate}  following MagicHOI’s setting.}
    \vspace{-3mm}
    \label{fig:qual_magichoi}
\end{figure}
\begin{table}[t]
\small
\caption{\small Results on HO3D-v3~\citep{hampali2020honnotate} following MagicHOI's setting.}
\label{tab:ho3d_magichoi}
\centering  
    \begin{tabular}{lcccccc}
    \toprule
    \multirow{2.5}{*}{Method}    & Hand  & \multicolumn{4}{c}{Object} & Interaction \\ 
    \cmidrule(r){2-2} \cmidrule(r){3-6} \cmidrule(r){7-7}
    & MPJPE $\downarrow$ & CD  $\downarrow$ & F5 $\uparrow$ & F10 $\uparrow$ & RS $\downarrow$ & CD$_h$ $\downarrow$ \\ 
    \midrule
    iHOI      &  27.75   & 2.37  & 35.78 & 62.11  & 0.17  &   25.45   \\ 
    DiffHOI   &  16.02   & 2.30  & 39.59 & 64.49  & \cellcolor{third}0.13  &   33.33   \\
    EasyHOI   &  16.69   & 1.86  & 46.10 & 70.92  & 0.28  &   19.55   \\
    HOLD      &  30.79   & 1.31  & 57.20 & 80.23  & 0.62  &   21.28   \\
    MagicHOI  &   \cellcolor{second}4.62   & \cellcolor{third}0.87  & \cellcolor{third}69.72 & \cellcolor{third}92.15  & \cellcolor{second}0.11  &    \cellcolor{second}2.39   \\
    CHOIR     &   \cellcolor{third}5.55   & \cellcolor{second}0.77  & \cellcolor{second}72.33 & \cellcolor{second}96.03  & \cellcolor{best}\textbf{0.10}  &    \cellcolor{third}2.41   \\  
    \textbf{\ourmethod (ours)}     & \cellcolor{best}\textbf{4.60}   & \cellcolor{best}\textbf{0.24}  & \cellcolor{best}\textbf{90.24} & \cellcolor{best}\textbf{98.13}  & \cellcolor{third}0.13   & \cellcolor{best}\textbf{2.17}      \\ 
    \bottomrule
    \end{tabular}
\end{table}

\section{More Experimental Results}
\subsection{HO3D-v3 MagicHOI's setting}
We empirically observe that some methods perform better on shorter clips, motivating an additional evaluation using the same sequences and frames as MagicHOI~\citep{wang2025magichoi}.
Table~\ref{tab:ho3d_magichoi} and Figure~\ref{fig:qual_magichoi} present the quantitative and qualitative results, respectively.
Under this protocol, \ourmethod achieves the best results on five of the six metrics, with substantial improvements in object reconstruction accuracy.

\subsection{Comparison of Object Generation}
\begin{figure}[t]
    \centering
    \small
    \includegraphics[width=1\linewidth]{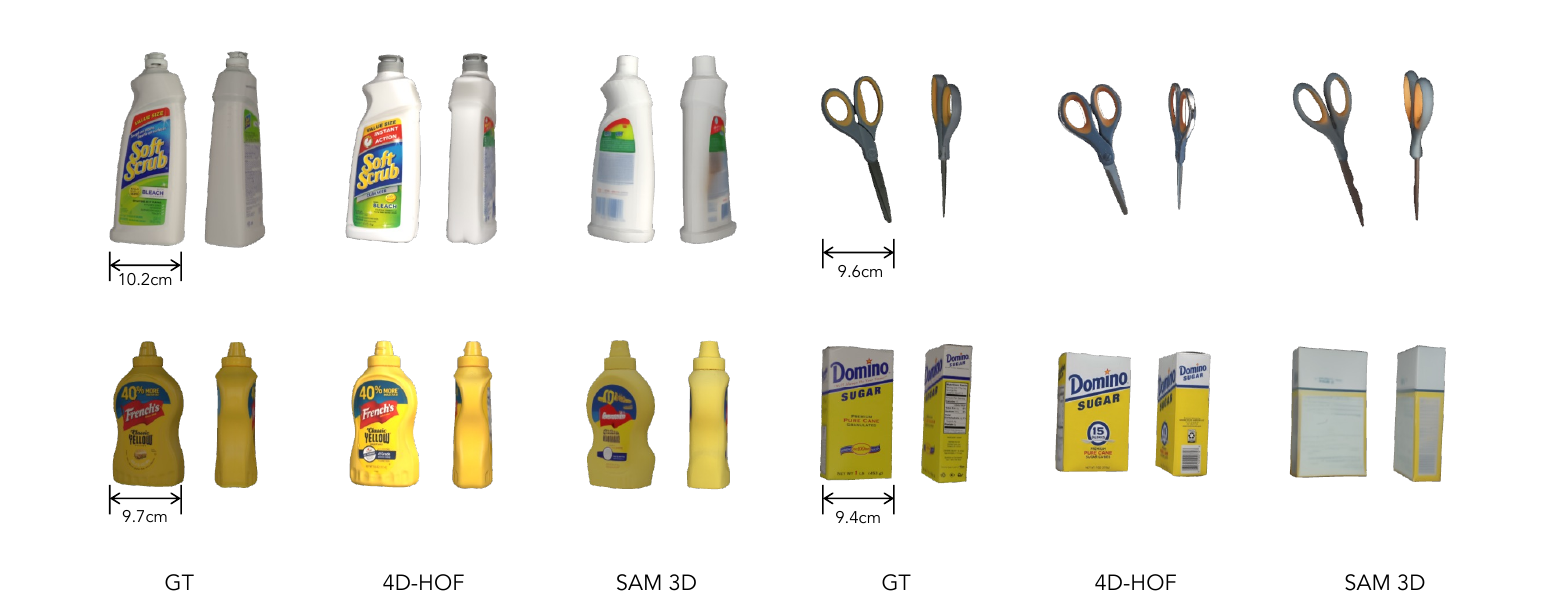}
    \caption{\small Qualitative comparison of object generation. We show two views of each object reconstructed by \ourmethod and SAM3D~\citep{chen2026sam} alongside the ground truth. Our pipeline better preserves object geometry and texture details, including labels and fine structures. Ground-truth object widths are provided for scale reference.}
    \label{fig:object}
\end{figure}

We compare our multiview-based object generation pipeline with SAM3D~\citep{chen2026sam}, with quantitative and qualitative results presented in Table~\ref{tab:object} and Figure~\ref{fig:object}, respectively. Limited to a single input image, SAM3D produces less reliable textures, which can compromise downstream pose estimation with FoundationPose, where rendered object appearance provides a key cue for pose evaluation. By combining complementary observations from multiple views with a powerful image generation model, our pipeline produces objects with higher geometric accuracy and texture fidelity.
\begin{table}[t]
\small
\caption{\small Object generation comparison. Our multiview-based pipeline outperforms SAM3D across all metrics, demonstrating the benefit of complementary visual observations for object reconstruction.}
\label{tab:object}
\centering
    \begin{tabular}{lcccc}
    \toprule
    Methods & CD$\downarrow$ & F5$\uparrow$ & F10$\uparrow$ & RS$\downarrow$ \\ 
    \midrule
    SAM3D & 0.47 & 77.54 & 96.20 & 0.17 \\ 
    \ourmethod (Ours) & 0.28 & 89.11 & 97.13 & 0.12 \\ 
    \bottomrule
    \end{tabular}
\end{table}

\subsection{Do more HOI data help?}
\begin{figure}[t]
    \centering
    \small
    \includegraphics[width=0.4\linewidth]{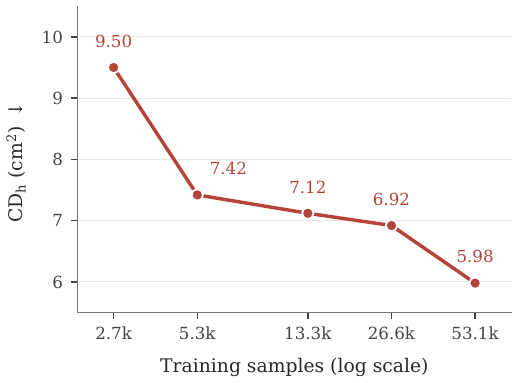}
    \caption{\small Dataset scaling analysis for bridge matching model training. }
    \label{fig:scaling}
\end{figure}

To investigate the impact of training data volume, we evaluate our model's performance on the HO3D dataset using the CD$_h$ metric (Figure \ref{fig:scaling}). The results indicate that increasing the size of the training dataset yields consistent performance gains, demonstrating the scalability of our approach.

\subsection{Comparison with GPT 6 Astra Ultra (as of Sep 20, 2026)}
\label{ss:gpt6}
\begin{figure}[t]
    \centering
    \small
    \includegraphics[width=0.98\linewidth]{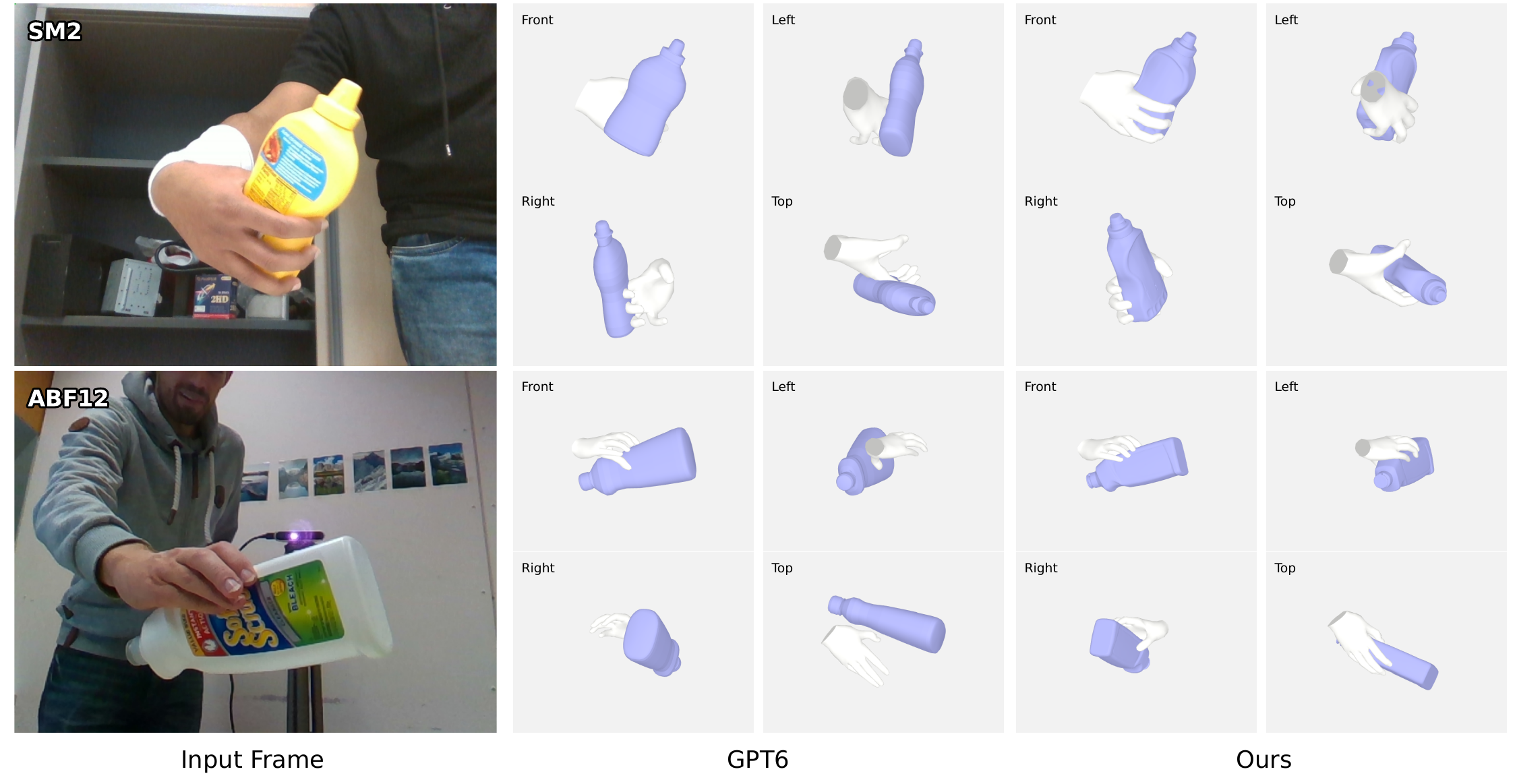}
    \caption{\small {Qualitative comparison between GPT6-Astra and our 4D-HOF}.}
    \label{fig:gpt6_compare}
\end{figure}

Recent GPT-6~\citep{openai2026chatgpt} models have demonstrated strong capabilities in 3D reasoning and tool use. To provide a broader and more rigorous evaluation of our model, we constructed an agentic baseline by prompting GPT-6-Astra with Ultra reasoning effort and providing access to FoundationPose~\citep{wen2024foundationpose}, WiLoR~\citep{potamias2025wilor}, MoGe-3~\citep{kong2026moge}, Blender~\citep{blender}, and related tools. We tasked the agent with reconstructing 4D hand–object interactions from the ABF12 and SM2 sequences of HO3D dataset , without imposing execution-time or token-budget limits. Our prompt is illustrated in~\ref{box:gpt6_prompt}. Qualitative comparisons are shown in Figure~\ref{fig:gpt6_compare}. Although GPT-6 recovers plausible coarse hand and object poses, it exhibits noticeable hand–object separation and weaker cross-view contact consistency. In contrast, our method better preserves instance-specific object geometry and produces tighter, more physically plausible hand–object configurations: in SM2, the fingers more consistently wrap around the bottle, while in ABF12, our reconstruction better captures the complex fingertip contact and provide consistent result across viewpoints. These results suggest that, despite GPT-6’s strong general-purpose 3D reasoning ability, the proposed pipeline provides more accurate geometry and more coherent hand–object interactions.

Beyond reconstruction quality, the agentic baseline incurs substantial runtime and resource overhead. In our experiments, a single run typically required several hours and consumed approximately 35\% of the weekly usage allowance associated with our \$100 subscription plan, making large-scale evaluation impractical under this budget. These observations highlight the challenges of applying general-purpose agentic systems to efficient, contact-consistent HOI reconstruction. Our feed-forward module could serve as a specialized component within such systems to accelerate reconstruction while improving geometric and interaction consistency.

\begin{tcolorbox}
[center,colback=gray!5!white,colframe=gray!75!black,title=GPT6 HOI Reconstruction Prompt]
Two third-person-view human hand–object interaction videos is available at: data/ABF12.mp4, data/SM2.mp4 \\

For each video, reconstruct:\\
1. A temporally consistent MANO hand sequence.\\
2. A textured mesh for each manipulated object.\\
3. A per-frame 6DoF object pose sequence.\\

The hand and object should share a consistent metric coordinate system. Minimize hand–object penetration and separation, keep motion smooth, preserve plausible physical contact, and ensure rendered hand/object meshes align closely with the original RGB frames in 2D. Handle occlusion and tracking loss without large pose jumps, scale changes, or identity switches.\\

If additional Python packages are needed, create a virtual environment in the current directory rather than modifying the system environment. Blender and Blender-mcp are installed and enabled.\\

- Use MoGe-3 (https://github.com/microsoft/MoGe) for monocular depth estimation. \\
- Use WiLoR (https://github.com/rolpotamias/WiLoR) for MANO hand reconstruction, plus additional optimization procedures. \\
- Use FoundationPose (https://github.com/NVlabs/FoundationPose) for object 6DoF pose estimation, and make suitable modifications if you need. Create textured mesh and adjust mesh scale before running the model. You can use the depth from MoGe 3 as input.\\

Output:\\
- MANO parameters, joints, and vertices.\\
- Textured object meshes.\\
- Per-frame object poses and confidence.\\
- Camera parameters and coordinate-system documentation.\\
- RGB overlay videos and necessary 3D/diagnostic visualizations.\\
- README and other documents containing setup, commands, methods, optimization details, output paths, validation results, and known limitations.
\end{tcolorbox}
\label{box:gpt6_prompt}

\section{Failure Case Analysis}
\begin{figure}[t]
    \centering
    \includegraphics[width=0.8\textwidth]{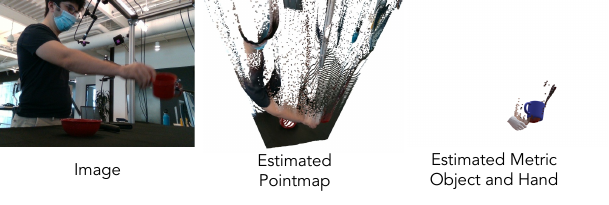}
    \caption{\small Failure analysis. When the estimated point map contains severe artifacts or out-of-distribution predictions,
the resulting initialization may become unreliable, which can affect the flow-matching refinement process and lead to reconstruction failures.}
    \label{fig:failure}
\end{figure}

Our pipeline is reliant on several vision foundation models, with the monocular depth estimator, MoGe-3, acting as a critical geometric anchor. MoGe-3~\citep{kong2026moge} recovers the object's metric scale to match the hand and coarsely aligns both components within a unified coordinate space. Consequently, if the predicted point map exhibits significant artifacts (Figure ~\ref{fig:failure}), these inaccuracies cascade directly into the object scale and pose estimation modules. Such highly corrupted, out-of-distribution initializations ultimately misguide the flow-matching refinement model, resulting in catastrophic reconstruction failures. 

\section{VLM Prompt}

Our hand and object states initialization pipeline employs VLM for multiple purposes, including identifying the manipulated object, generating multi-view images, and selecting object textures. We describe each task below and provide the corresponding prompts.

\subsection{Interaction object reasoning}
 Given an input video, the VLM identifies the object manipulated by the hand and generates a textual description. This description is then passed to SAM3 as a prompt for automatic object segmentation.

\begin{tcolorbox}
[center,colback=gray!5!white,colframe=gray!75!black,title=Object Identification Prompt]
Describe the object interaction with hand, only return in [color object\_name] format.
\end{tcolorbox}
\label{box:object}

\subsection{Multi-view image generation}
We select object keyframes from the input video and provide them to an image generation model to synthesize canonical views of the object. The placeholder \texttt{\{description\}} in the prompt is filled with the object description obtained in the preceding reasoning step. 
\begin{tcolorbox}
[center,colback=gray!5!white,colframe=gray!75!black,title=Canonical View Generation Prompt]
Generate the front, back, left, right, top, bottom canonical views of this \{description\}. The generated views MUST perfectly match the exact shape, texture, color, and geometric details of the object in the reference images. Complete the occluded regions. Present as a 2*3 grid, front, back, left in the upper row, right, top, bottom in the lower row. Pure white background. The image must not contain text indicating viewpoint direction.
\end{tcolorbox}
\label{box:multiview}

After each generation attempt, the VLM acts as a critic to assess whether the generated image satisfies the specified requirements. If the image fails this evaluation, we regenerate it. 
\begin{tcolorbox}
[center,colback=gray!5!white,colframe=gray!75!black,title=Generated View Evaluation Prompt]
You are an expert evaluator. Look at the provided generated image at the very end.
\\
It is supposed to be a 2x3 image grid showing the front, back, left, right, top, and bottom canonical views of the object in the reference images.
\\
Does the generated image strictly follow these requirements: 

1. It is exactly a 2x3 grid. 

2. Background is pure white. 

3. No text indicating viewpoint direction. 

4. Identity and texture are preserved.
\\
Answer EXACTLY 'Yes' or 'No' without any explanation.
\end{tcolorbox}
\label{tab:eval}

If both consecutive attempts fail, the VLM selects the better of the two candidates, allowing the pipeline to proceed.
\begin{tcolorbox}
[center,colback=gray!5!white,colframe=gray!75!black,title=Candidate Image Selection Prompt]
I have two generated images at the end, each supposed to be a 2x3 grid showing canonical views of the same object from the preceding reference images. Image 1 is the previous best. Image 2 is the newly generated one. 
\\
Compare them based on these criteria: 

1. Exactly a 2x3 grid layout. 

2. Pure white background.

3. No text or labels. 

4. High visual quality and accurate canonical views.
\\
Which image strictly follows these instructions better and has better quality?

Return EXACTLY 'Image 1' or 'Image 2' without any explanation or markdown formatting.
\end{tcolorbox}
\label{box:select}

\subsection{Texture selection} 
Given the reconstructed object geometry, we obtain two candidate texture maps through direct generation and image projection, respectively. We apply each texture map to the geometry and render the resulting textured object from multiple viewpoints. The VLM then compares these multi-view renderings and selects the better texture.
\begin{tcolorbox}
[center,colback=gray!5!white,colframe=gray!75!black,title=Texture Selection Prompt]
I have two 2x3 image grids showing canonical views of the same 3D object's geometry. Image 1 represents Model A. Image 2 represents Model B. Important: The 3D geometry is identical in both models, but the poses or viewpoints might differ between the two images.

Please act as a 3D texture quality judge. Ignore any pose or orientation differences, and focus strictly on comparing the texture quality.
Look for detail, sharpness, realism, and a lack of noise or artifacts. Which model has better texture quality? 

Return EXACTLY 'Model A' or 'Model B' without any explanation or markdown formatting.
\end{tcolorbox}
\label{box:texture}

\end{document}